\documentclass[letterpaper, 10 pt, conference]{ieeeconf}  

\usepackage{subfigure}
\usepackage{float}
\usepackage{mathptmx} 
\usepackage{amsmath} 
\usepackage{amssymb}  
\usepackage[noend]{algpseudocode}
\usepackage{booktabs}
\usepackage{cite}
\usepackage{multicol}
\usepackage{multirow}
\usepackage{bbding}
\usepackage{booktabs,makecell,indentfirst,epsfig}
\usepackage {multicol,multirow,color}
  \usepackage{array}
\usepackage[caption=false,font=normalsize,labelfont=sf,textfont=sf]{subfig}
  \usepackage{textcomp}
  \usepackage{stfloats}
  \usepackage{url}
  \usepackage{verbatim}
  \usepackage{graphicx}
  \usepackage{cite}
  \usepackage{bbm}
  \usepackage{dsfont}

\IEEEoverridecommandlockouts                              

\title{\LARGE \bf
SLTNet: Efficient Event-based Semantic Segmentation with Spike-driven Lightweight Transformer-based Networks
}

\author{Xianlei Long$^1$, Xiaxin Zhu$^1$, Fangming Guo$^1$, Wanyi Zhang$^1$, Qingyi Gu$^2$, Chao Chen$^1$, and Fuqiang Gu$^1$$^*$
\thanks{This work is supported by China Postdoctoral Science Foundation (No. 2023M740402), in part by the Fundamental Research Funds for the Central Universities (No. 2024CDJGF-034, 2024CDJGF-049), in part by National Natural Science Foundation of China (No. 62403085, No. 42174050, No. 42474027), and Chong Startup Project for Doctorate Scholars (No. CSTB2022BSXM-JSX005). \emph{Corresponding: Fuqiang Gu.} }
\thanks{$^1$ Xianlei Long, Xiaxin Zhu, Fangming Guo, Wanyi Zhang,  Chao Chen, and Fuqiang Gu are with the College of Computer Science, Chongqing University, Chongqing 400044, China
         \tt\small (e-mails: xianlei.long@cqu.edu.cn, 202214021084t@stu.cqu.edu.cn, \{202114131181, wanyi.zhang, cschaochen, gufq\}@cqu.edu.cn).}%
         \thanks{$^2$ Qingyi Gu is with the Institute of Automation, Chinese Academy of Sciences, Beijing 100190, China \tt\small (e-mail: qingyi.gu@ia.ac.cn).}
}

\begin{document}

\maketitle
\thispagestyle{empty}
\pagestyle{empty}

\begin{abstract}
Event-based semantic segmentation has great potential in autonomous driving and robotics due to the advantages of event cameras, such as high dynamic range, low latency, and low power cost. Unfortunately, current artificial neural network (ANN)-based segmentation methods suffer from high computational demands, the requirements for image frames, and massive energy consumption, limiting their efficiency and application on resource-constrained edge/mobile platforms. To address these problems, we introduce SLTNet, a Spike-driven Lightweight Transformer-based Network designed for event-based semantic segmentation.
Specifically, SLTNet is built on efficient spike-driven convolution blocks (SCBs) to extract rich semantic features while reducing the model's parameters. Then, to enhance the long-range contextual feature interaction, we propose novel spike-driven transformer blocks (STBs) with binary mask operations. Based on these basic blocks, SLTNet employs a high-efficiency single-branch architecture while maintaining the low energy consumption of the Spiking Neural Network (SNN).
Finally, extensive experiments on DDD17 and DSEC-Semantic datasets demonstrate that SLTNet outperforms state-of-the-art (SOTA) SNN-based methods by at most 9.06\% and 9.39\% mIoU, respectively, with extremely 4.58$\times$ lower energy consumption and 114 FPS inference speed. Our code is open-sourced and available at \url{https://github.com/longxianlei/SLTNet-v1.0}.

\end{abstract}

\section{Introduction}

Semantic segmentation \cite{ss} aims to divide visual data into distinct regions with clear semantic properties, thereby enabling a deep understanding of the scene. This technology plays a crucial role in various cutting-edge fields such as autonomous driving~
\cite{driving}, security monitoring~\cite{long_scene_segmentation}, and robotics, forming the foundational component for environmental perception of intelligent robotics. 
However, with the increasing efficiency demands for robotics in complex scenarios, traditional image-based semantic segmentation methods show significant performance degradation when processing high-dynamic or variable lighting scenes~\cite{Ni_Conext_seg, Ni_ssaseg}. 
This challenge has driven the development of event camera-based applications.


Event cameras, leveraging their innovative biomimetic event generation characteristics, have inspired considerable research interests in computer vision and robotics communities~\cite{event_research}. 
Event cameras excel at detecting variations in light intensity at the pixel level with remarkable microsecond resolution. 
The event-driven nature of these cameras shows several superior advantages, including high dynamic range (HDR) perception, low latency, and low energy cost, showing promising solutions to computationally intensive semantic segmentation~\cite{Gu_eventdrop, spike-brgnet}.
Although research on event-based semantic segmentation is still in its infancy, several works have been proposed.
Ev-SegNet~\cite{evsegent} develops the first baseline using solely events but requires a substantial computational load. Following this, DTL \cite{dual}, Evdistill \cite{evdistill}, and ESS \cite{ess} attempt to compensate for the lack of visual details in event data through transfer learning or knowledge distillation technologies from image processing branches.
However, existing methods remain computationally expensive or rely on auxiliary images~\cite{yao_sam_segmentation}. Therefore, a critical challenge is to design an efficient and robust event-based segmentation model that minimizes computational and energy costs while boosting event feature extraction ability.




With the development of biological computing, Spiking Neural Network (SNNs) is a potential solution inspired by mechanisms of information generation and transmission in the brain, which can convey dynamically generated binary spikes across spatial and temporal dimensions~\cite{snn}. Thus, SNNs demonstrate excellent advantages in computation efficiency, temporal memory capacity, and biological interpretability. 
The basic building block of SNNs is the spiking neuron, which shares a highly similar generation mechanism with the event cameras, making SNNs theoretically well-suited to work with such cameras. Encouraged by these advantages, some works~\cite{kim2022beyond,evsegsnn} attempt to integrate SNNs with classical ANN-based segmentation networks, but the performance is poor due to the plain feature extraction module. This indicates that combining the SNNs with event segmentation still requires further exploration.

To address these issues, we propose SLTNet, a Spike-driven Lightweight Transformer-based semantic segmentation network that utilizes events only to achieve energy-efficient and robust performance.
SLTNet is a hierarchical single-branch SNN-based method equipped with an encoder-decoder structure. It can extract both detailed and semantic information from only event data while maintaining high efficiency and low energy consumption. Specifically,
SLTNet contains four stages in the encoder part, which include three novel Spike-driven Convolution Blocks (SCBs) that capture fine-grained textural information, and two newly designed Spike-driven Transformer Blocks (STBs) that are responsible for extracting long-range contextual features in the last stage.
Then, to further reduce the computational cost, we propose a Spiking Lightweight Dilated module (Spike-LD) that serves as the basic block of our conv-based SCBs and decoders.
Finally, extensive experimental results conducted on two recognized public event segmentation datasets demonstrate that the SLTNet outperforms SNN-based SOTA methods by a large margin while maintaining extremely low energy consumption with faster inference speed.



Main contributions of this work can be summarized as:
\begin{itemize}
\item We propose a novel SLTNet segmentation model with events only to deal with high-dynamic complex scenarios, which is constructed on the basic SCBs and STBs that enable high-efficiency feature extraction and low computation cost.

\item  To enhance feature interaction, we propose a novel STB module by leveraging the re-parameterization convolution and spike-driven multi-head self-attention mechanism to achieve long-range contextual feature interaction with only floating-point accumulation operations.

\item To reduce model parameters while maintaining efficient feature extraction ability, we design a Spike-LD module that can capture multi-scale event features concurrently and can easily adapt to SNN architectures.



\item {Extensive experimental results show that SLTNet outperforms SNN-based methods by at most 9.06\% and 9.39\% mIoU with much lower computation cost and 114 FPS inference speed.}

\end{itemize}

\section{Related Works}
\subsection{Event-based Semantic Segmentation}

Event cameras feature a unique event-driven mechanism that makes them highly sensitive to moving objects. This high sensitivity presents a challenge in developing novel algorithms to extract meaningful information from event data effectively. Recently, event camera-based applications have become a prominent research area, including their application in semantic segmentation~\cite{spike-brgnet,EventAug_tcds}.
Ev-SegNet~\cite{evsegent} introduces the first baseline for semantic segmentation using only event data, leveraging an Xception-like model and an automatic labeling method to generate approximate semantic annotations.
However, the sparse nature of event streams and the limited high-quality labeled datasets hinder the development of event-based methods to surpass image-based approaches. To address these, recent efforts focus on transferring knowledge from images to enhance performance. E.g., EvDistill~\cite{evdistill} employs knowledge distillation to incorporate features from a teacher network trained on large-scale labeled image data, achieving remarkable performance on unlabeled event data. Similarly, ESS~\cite{ess} utilizes unsupervised domain adaptation to transfer semantic segmentation tasks from labeled image datasets to unlabeled event data.


However, existing studies adopt processing paradigms originally designed for images, neglecting the inherent spatiotemporal relationships within event streams. As a result, event-based semantic segmentation algorithms remain underdeveloped, highlighting the need for our proposed approach.

\subsection{SNN-based Semantic Segmentation}
{To tackle the above problems, researchers resort to the SNN for its superior performance in energy efficiency, temporal memory capacity, and biological interpretability. Kim et al.~\cite{kim2022beyond} and EvSegSNN~\cite{evsegsnn} incorporate spiking neurons into classical FCN, DeepLab, and U-Net segmentation architectures, but these models showed unsatisfactory performance. Then, MFS-EFS~\cite{mfs-efs} proposes a spiking context-guided architecture with biologically inspired feature extraction blocks, enabling unified semantic segmentation across frame and event-based vision modalities through direct spike-driven training. 
To improve performance, SCGNet~\cite{SCGNet} proposes a multi-scale fully spiking segmentation network with depthwise-separable residual blocks, pioneering the use of surrogate gradient training for deep SNN-based semantic segmentation. Spike-BRGNet~\cite{spike-brgnet} adopts a three-branch SNN design to balance accuracy and energy consumption, though its computational cost remains high.

In summary, existing approaches still exhibit critical deficiencies in both energy efficiency and computational performance. Designing an efficient and compact architecture remains a challenge. This work aims to achieve an optimal balance by introducing a novel SNN architecture specifically designed for event stream segmentation.
}

\section{Fundamentals of SNNs and ESNs}
\label{sec: blocks design}
\begin{figure}[!t]
    \centering 
\includegraphics[width=0.70\linewidth]{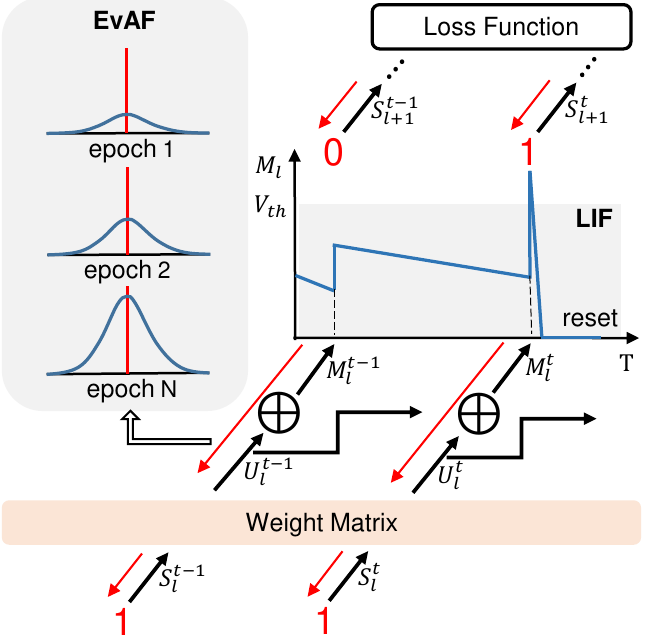}	
    \caption{The workflow of the evolutionary spiking neuron model in our algorithm. The black line denotes the forward propagation route, while the red line indicates the backward propagation process.} 
    \label{Fig:LIF}
\end{figure}
To construct a robust evolutionary spiking neuron (ESN) in SLTNet, we employ the Leaky Integrate-and-Fire (LIF) as the basic neuron due to its excellent biological interpretability and computational efficiency. We show its workflow in Fig.~\ref{Fig:LIF}. Mathematically, we explicitly formulate ESN's iterative forward propagation process as follows:
\begin{equation}
\begin{gathered}
M^t_l=\tau U^{t-1}_l+W^t_lX^t_{l-1},\\
S^t_l=H(M^t_l-U_{th}), \\
U^t_l=U_{reset}S^t_l+U^t_l(1-S^t_l),
\end{gathered}
\label{eq:seg}
\end{equation}
where the $U^t_l$ is the membrane potential of neurons in the $l^{th}$ layer at time step $t$, $W^t_l$ is synaptic weight matrix between layers $l-1$ and $l$, $S^t_l$ is the output of current layer, and $\tau$ is membrane time constant. 
The LIF module receives the leaky membrane potential of the previous time step $\tau U^{t-1}_l$ and calculates the current potential $W^t_lX^t_{l-1}$. When the accumulation reaches threshold $U_{th}$, the neuron fires, generating a spike $S^t_l$ using Heaviside step function $H(\cdot)$; otherwise, it remains zero. After firing, the membrane potential rebounds to $U_{reset}$.

We train our model with an end-to-end strategy, so the back-propagation in the ESN module can be described as:
\begin{equation}
\label{eq:Backpropagation in SNN}
    \frac{\partial L}{\partial W^t}=
    \sum_t(\frac{\partial L}{\partial S^t}\frac{\partial S^t}{\partial U^t}
          +\frac{\partial L}{\partial U^{t+1}}\frac{\partial U^{t+1}}{\partial U^t})
          \frac{\partial U^t}{\partial W^t},
\end{equation}
where $L$ represents the loss function, $W^t$ denotes the weights, and $S^t$ and $U^t$ indicate the activated spike and membrane potential, respectively. 

The gradient of the spiking activity function ${S^t}/{U^t}$ signifies their non-differentiable nature: either at a standstill or escalating to infinite magnitudes. We solve this problem by utilizing the evolutionary surrogate gradient function, called EvAF \cite{evaf}. It will dynamically adapt to generate substitute gradient values as epochs iterate:
\begin{equation}
\begin{gathered}
\varphi(x)=\frac{1}{2}\tanh{K(i)(x-U_{th})}+\frac{1}{2},\\
K(i)=\frac{(10^{\frac{i}{N}}-1)K_{max}+(10-10^{\frac{i}{N}})K_{min}}{9}, 
\end{gathered}
\label{eq:EvAF}
\end{equation}
where $i\in [0,N-1]$ is the index of training epoch.
Following the common setup in \cite{evaf}, we set $K_{min}$ = $1$ and $K_{max}$ = $10$.

\section{Overall Architecture of SLTNet}



We first introduce the event representation method, which converts raw event streams into formats suitable for network processing. Next, we present the efficient Spike-LD and STB modules, which are the fundamental building blocks of SLTNet. Finally, we describe the overall SLTNet architecture and its segmentation loss.


\subsection{Event Representation}
We encode the raw asynchronous event stream $e_i=\{(x_i,y_i,t_i,p_i)\}_{i\in \Delta t}$ into voxel grid $V_{k,x,y}\in K\times H\times W$:
\begin{equation}
    V_{k,x,y} = \sum_{i\in \Delta t}\delta(\lfloor \frac{t_i}{\Delta t}\rfloor=k)\delta(x_i = x)\delta(y_i=y)p_i, 
\end{equation}
where $\Delta t$ is time interval, time index $k=\lfloor\frac{t_i}{\Delta t}\rfloor$, $\delta$ denotes Kronecker Delta function used for indicating whether event $e_i$ is within time period $k$ and located at $(x, y)$, $p_i\in \{-1,+1\}$ represents the polarity. 
Furthermore, to adapt the spiking paradigm, we transform the voxel grid $V_{k,x,y}$ into an event tensor, formatted as $E(x,y,t)\in T\times K\times H\times H$ that SNNs could calculate through the temporal dimension. $E(x,y,t)$ will be fed into our SLTNet as event representations.

\subsection{Spiking Lightweight Dilated Module}

\begin{figure}[!t]
    \centering 
\includegraphics[width=0.8\linewidth]{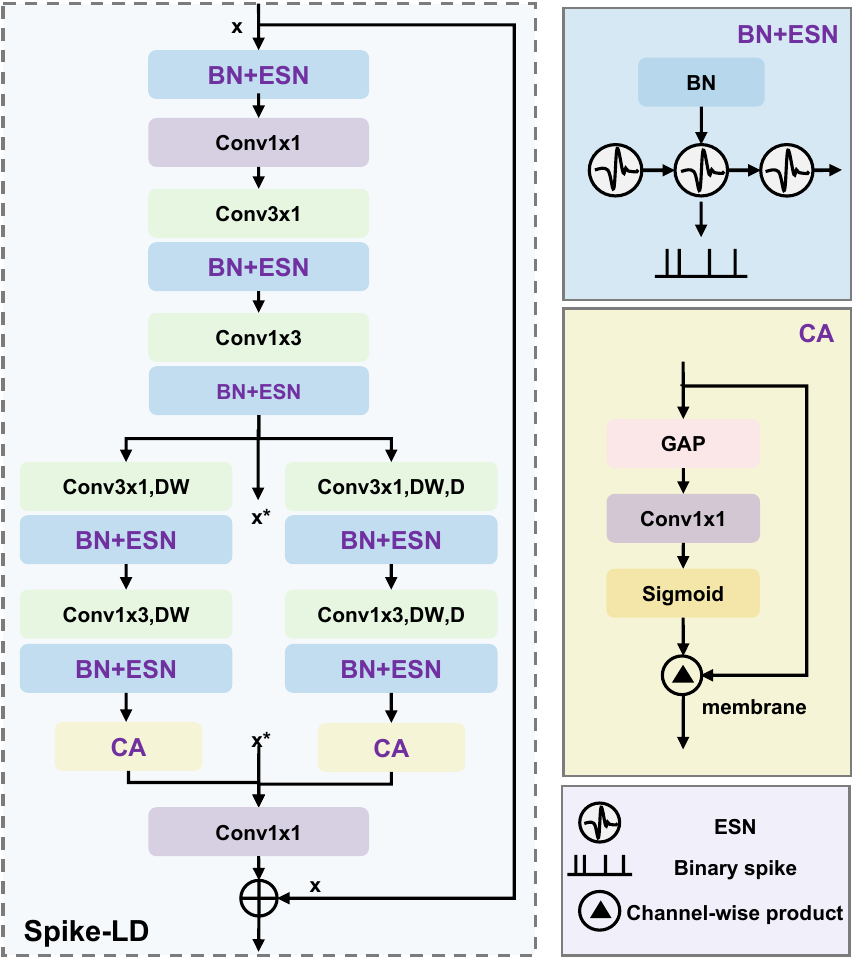}	
    \caption{The detailed construction of Spike-LD module. Among them, $DW$ is depth-wise convolution, $D$ denotes dilated convolution, $GAP$ means global average pooling and $membrane$ represents the membrane potential within the spiking neuron. ESN is an Evolutionary Spiking Neuron.} 
    \label{Fig:Spike_LD}
\end{figure}

    

As illustrated in Fig.~\ref{Fig:Spike_LD}, the Spike-LD module adopts a bottleneck structure~\cite{resnet} with a membrane shortcut, which establishes a shortcut connection between the membrane potential of spiking neurons, addressing the vanishing gradient problem by achieving identity mapping in a spike-driven mode. Then, it integrates the spiking paradigm by following the weight computation stage with batch normalization (BN) and the ESN model introduced previously.

In the feature extraction stage, considering the parameter increase that traditional $3\times3$ convolutions might cause, the module utilizes decomposed convolution, a combination of $1\times3$ and $3\times1$ convolutions. This allows the network to independently extract features from both the width and height directions while maintaining sensitivity to local features without increasing the parameters. Considering the importance of multi-scale features in event semantic segmentation tasks, our Spike-LD module introduces a three-branch structure, combining dilated and depth-wise convolution to capture features within different receptive fields. Moreover, to compensate for the loss of inter-channel interaction information caused by depth-wise convolution, the module employs channel attention (CA) modules~\cite{CA} to improve the representation of features.
Finally, $1\times1$ convolutions at the beginning and end of the module serve to compress and expand channel dimensions, respectively, to reduce the computational load of the model.

\subsection{Spike-Driven Transformer Block}

\begin{figure*}[t]
    \centering 
\includegraphics[width=0.95\linewidth]{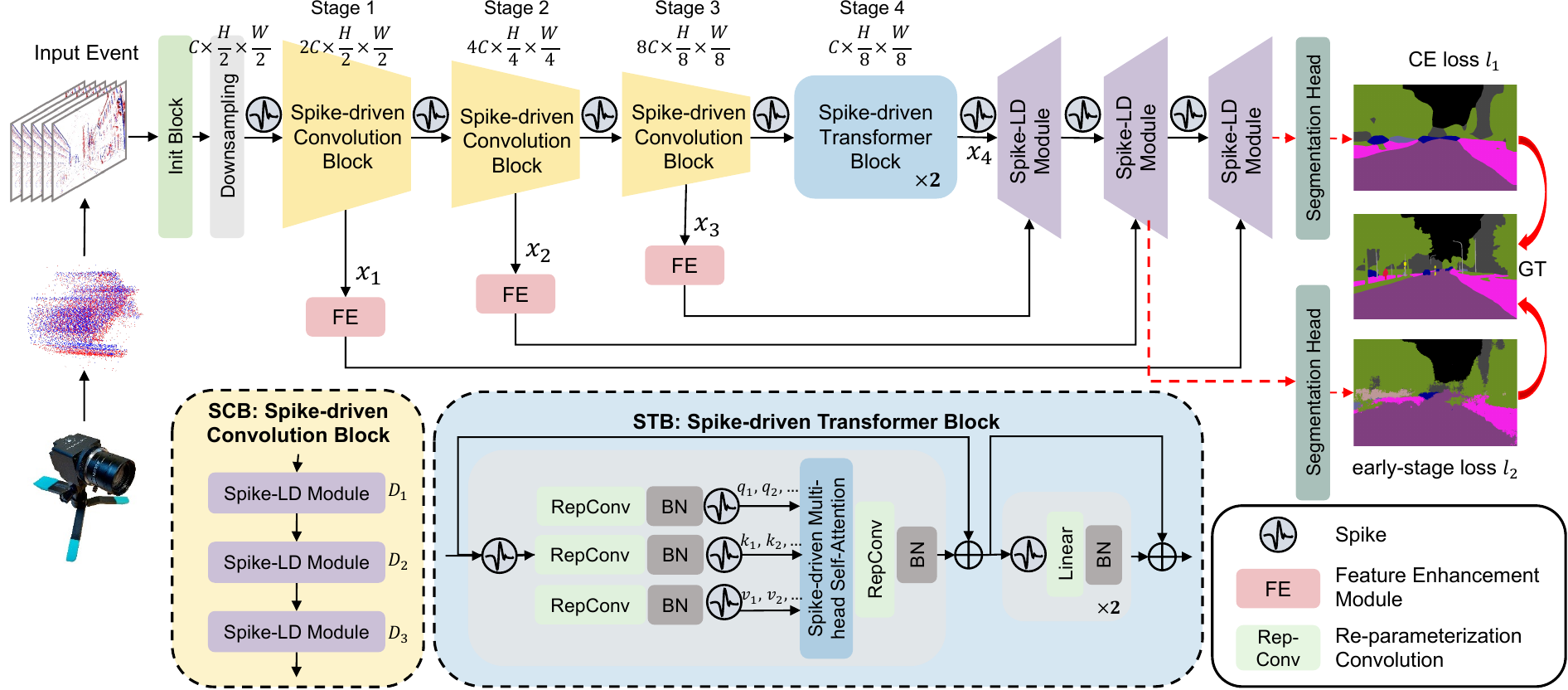}	
    \caption{The architecture of SLTNet, where input events are passed through a four-stage spiking encoder built on SCBs and STBs. Then, features are enhanced by three Spike-LD modules in the decoder. Finally, two separate segmentation heads generate two prediction probability maps at different levels.}
    \label{Fig:architecture}
\end{figure*}

Inspired by Yao et al. \cite{yao2024spikedrivenv2}, we design an STB as shown at the bottom of Fig.~\ref{Fig:architecture}. It comprises a spike-driven multi-head self-attention module (SDMSA) aimed at capturing spatial relationships and a two-layer MLP complementing channel information. The main process of STB is formulated as:
\begin{equation}
\begin{gathered}
Q ,K,V= SN[RepConv_1(M),RepConv_2(M),RepConv_3(M)]
\end{gathered}
\end{equation}
\begin{equation}
\label{eq:SDMSA}
    SDMSA(Q,K,V) = SUM_c(Q\otimes K)\otimes V,
\end{equation}
\begin{equation}
\begin{gathered}
    M' = M + RepConv_4(SDMSA(Q,K,V)),\\
    MLP(M') = Linear_2(SN_2(Linear_1(SN_1(M')))),\\
    M'' = M' + MLP(M'),
\end{gathered}
\end{equation}
where SN, RepConv, and $\otimes$ are spiking neurons, re-parameterization convolution~\cite{repconv}, and Hadamard product, respectively. We omit the BN operation for clarity.

The design of residual connections in spike transformers is critical; unlike the vanilla shortcut and the spike-element-wise shortcut~\cite{sew_shortcut}, we use the membrane shortcut, which performs shortcuts between membrane potentials to enhance performance while maintaining spiking features.


STB incorporates self-attention in a spiking-friendly manner and offers two key advantages over traditional CNN–Transformer blocks. First, as shown in Eq.~\ref{eq:SDMSA}, our SDMSA mechanism reduces computational complexity.
This makes the block far more scalable to high-resolution feature maps. Second, by leveraging spiking neurons, floating-point multiplications in the attention operation are replaced with masked operations that skip inactive spikes. As a result, the block primarily performs additions rather than multiplications, which substantially lowers energy consumption.
These design choices allow STB to retain the global context modeling capability of self-attention while maintaining the low-power, event-driven benefits of spiking neural networks.

\subsection{Spike-Driven Architecture of SLTNet}
\label{encoder-decoder}
As shown at the top of Fig.~\ref{Fig:architecture}, the preprocessed events $E(x,y,t)$ will pass through a hierarchical spiking encoder, a lightweight spiking decoder, and two segmentation heads, generating two prediction probability maps, $P_1$ and $P_2$. 

The hierarchical spiking encoder starts with an initial block, which stacks three spike convolution layers, each consisting of a $3\times3$ vanilla convolution, BN, and an ESN. Then, it goes through four stages: the first three stages are SCBs, each consisting of a downsampling layer and three spike-LD modules with different dilation rates, and the last stage includes two STBs. The SCB can extract local detail information, while the STB enriches the long-range context information that complements the local features. The downsampling layers applied in stages 1 to 3 consist of strided convolutions and pooling operations, transforming the feature dimensions from $C\times H\times W$ to $8C\times\frac{H}{8}\times\frac{W}{8}$. To reduce the computational cost, we narrow the feature channels to $C$ before feeding them into the STB. In addition, we place ESN between blocks to ensure spike transmission.

The lightweight spiking decoder is composed of three upsampling stages, each containing a Spike-LD module, an upsampling layer implemented using transposed convolution, BN, and a final pass through ESN neurons. To restore the spatial detail encoded in the earlier stage, we sequentially add the feature map generated by stages 1, 2, and 3 to different levels of decoders. Before fusion, the extracted event features $x_1$, $x_2$, and $x_3$ are enhanced by the feature enhancement (FE) module.
Finally, output features $F_1$ and $F_2$ are passed into two segmentation heads, generating two prediction probability maps, i.e., $P_1$ and $P_2$. 

\subsection{Segmentation Loss Design} As illustrated in Fig. \ref{Fig:architecture}, the designed loss consists of a cross-entropy (CE) semantic loss $l_1$ using Online Hard Example Mining (OHEM)~\cite{OHEM} and an early-stage CE semantic loss $l_2$.
Specifically, we use $l_1$ equipped with OHEM to improve the generalization ability by selecting the hardest samples from each mini-batch for training:
\begin{equation}
    l_1 = \sum_{i\in S}(-\sum_{j=1}^C y^{i}_j \log(\hat{y}^{i}_j)), \ l_2 = \sum_{i}(-\sum_{j=1}^C y^{i}_j \log(\hat{y}^{\text{early},i}_j)),
\end{equation}
where $S$ is the set containing the top $k$ percent of the largest losses among $l_i$. $l_2$ is employed to promote rapid convergence. It measures the discrepancy between the predicted probabilities, which are generated from the second-to-last Spike-LD module,  and the ground-truth label. 

Therefore, the entire loss in our SLTNet is: $Loss=\lambda_1l_1+\lambda_2l_2$. In our experiments, $\lambda_1$=1.0, $\lambda_2$=0.4, $k$=0.7.

\section{Experiments and Results}

\begin{table*}[!t]
    \centering
    \caption{Segmentation comparison of different ANN/SNN models on DDD17 and DSEC-Semantic dataset.}
    {
    \begin{tabular}{clccccccccc}
        \toprule 
        Dataset & Model & Data & Type & Params. (M) $\downarrow$ & FLOPs (G) $\downarrow$ & FPS $\uparrow$ & mIoU (\%) $\uparrow$\\
        \midrule
        \multirow{6}*{DDD17} & Ev-SegNet~\cite{evsegent} & Event & ANN & 23.75 & 44.17 & 20.67 & 54.81\\
        ~ & Evdistill~\cite{evdistill} & Event+Frame & ANN & 58.64 & 65.08 & 2.96 & 58.02\\
        ~ & ESS~\cite{ess} & Event+Frame & ANN & 6.69 & - & - & 61.37\\
        \cmidrule(lr){2-8}
        ~ & Spiking-DeepLab~\cite{kim2022beyond} & Event & SNN & 15.89 & 4.05 & 100 & 44.63 \\
        ~ & Spiking-FCN~\cite{kim2022beyond} & Event & SNN & 18.64 & 35.23 & 72.73 & 42.87\\

        ~ & {EvSegSNN}~\cite{evsegsnn} & Event & SNN & 8.55 & - & - & 45.54\\
        ~ & {SCGNet}~\cite{SCGNet}& Event & SNN & 0.49 & - & - & 49.30\\
        ~ & \textbf{SLTNet (Ours)} & \textbf{Event} & \textbf{SNN} & \textbf{0.41} & \textbf{1.96} & \textbf{114.29} & \textbf{51.93} \\
        \midrule
        \multirow{6}*{DSEC-Semantic} & Ev-SegNet~\cite{evsegent} & Event & ANN & 23.75 & - & - & 51.76\\
        ~ & ESS~\cite{ess} & Event+Frame & ANN & 6.69 & - & - & 51.57\\
        \cmidrule(lr){2-8}
        ~ & Spiking-DeepLab~\cite{kim2022beyond} & Event & SNN & 15.89 & 14.02 & 100 & 44.58\\
        ~ & Spiking-FCN~\cite{kim2022beyond} & Event & SNN & 18.64 & 126.46
 & 17.78 & 38.52\\
        ~ & {MFS-EFS}~\cite{mfs-efs} & Event & SNN & \textbf{1.11} & 28.88 & - & 46.10 \\
        ~ & \textbf{SLTNet (Ours)} & \textbf{Event} & \textbf{SNN} & {1.67} & \textbf{6.97} & \textbf{114.29} & \textbf{47.91} \\
        \bottomrule     
    \end{tabular}
    }
    \label{tab:seg_results}
\end{table*}

\subsection{Experimental Datasets}


Experiments are conducted on two widely used event-based semantic segmentation datasets: DDD17 and DSEC-Semantic (DSEC). The DDD17 dataset~\cite{ddd17} contains paired event data and grayscale images from driving scenes, captured with DAVIS sensors at $346\times 260$ resolution. Semantic labels, covering six categories, are generated using a pretrained segmentation model. The dataset includes 15,950 training and 3,890 testing pairs.
The DSEC dataset~\cite{dsec} provides event streams and high-resolution RGB images ($440\times 640$) from driving sequences, annotated with 11 fine-grained categories. Following the official split and preprocessing, it contains 8,082 training frames (eight sequences) and 2,809 testing frames (three sequences).

\subsection{Implementation Details}
Experiments are conducted on a single Nvidia GTX 4090 GPU with 24 GB of memory. We train our model using Adam Optimizer with StepLR strategy.
Training epochs, initial learning rate, weight decay, batch size, step size, and $\gamma$ in the StepLR strategy for DDD17 and DSEC-Semantic datasets are [300, 1e-3, 1e-4, 64, 5, 0.92] and [500, 1e-3, 1e-4, 16, 10, 0.92], respectively. For the LIF neuron in the ESN model, we set the reset value $U_{reset}=0$, the membrane time constant $\tau=0.25$, and the spike firing threshold $U_{th}=1.0$. In the event representation, the time interval $\Delta t=50$ ms and the membrane time constant $T=1$.

\subsection{Comparision with State-of-the-art Methods}

Given the limited number of studies on event-based semantic segmentation, our primary comparison involves several well-known methods, spanning two kinds of neural networks: ANNs and SNNs. For ANNs, we evaluate our approach against Ev-SegNet \cite{evsegent}, Evdistill \cite{evdistill}, and ESS \cite{ess}. In SNNs, we reimplemented Spiking-DeepLab and Spiking-FCN according to the network structure and implementation details described in \cite{kim2022beyond}. Due to unavailable code, we are unable to evaluate EvDistill on the DSEC-Semantic Dataset.

\subsubsection{Performance on DDD17}
Detailed results are given in Tab.~\ref{tab:seg_results}. In the field of SNNs, our SLTNet achieves the highest segmentation performance while maintaining the smallest number of parameters, minimal floating-point computations, and the fastest inference speed. Specifically, we achieve 7.30\%, 9.06\%, 6.39\% and 2.63\% higher mIoU than Spiking-DeepLab~\cite{kim2022beyond}, Spiking-FCN~\cite{kim2022beyond}, EvSegSNN~\cite{evsegent}, and SCGNet~\cite{SCGNet} respectively. Meanwhile, it has fewer parameters, 2.07$\times$ fewer FLOPs, and 1.14$\times$ faster FPS than comparative SNN methods. 

Next, we compare SLTNet to other ANN methods. Though our SLTNet slightly lags behind Ev-SegNet, it requires 57.93$\times$ fewer parameters, 22.54$\times$ less FLOPs, and achieves 5.53$\times$ faster FPS. Although there is a performance gap in mIoU compared to Evdistill and ESS, SLTNet achieves these results without using any image information and with significantly lower energy consumption and parameters. The energy consumption will be discussed in detail later.

\subsubsection{Performance on DSEC-Semantic}
As shown in Tab. \ref{tab:seg_results}, our model achieves the best segmentation result within the SNN field, with a performance of 47.91\% mIoU. Specifically, it outperforms Spiking-DeepLab by 3.33\% with 9.51$\times$ fewer parameters and Spiking-FCN by 9.39\% with 11.16$\times$ fewer parameters. {Furthermore, our method shows a 1.81\% performance enhancement over the latest MFS-EFS  while operating with 4.14$\times$ fewer FLOPs.} Though our model still falls behind ANN-based methods, the gap between them has significantly narrowed compared to the DDD17 dataset as the spatial resolution of the event data increases. 
At the same time, SLTNet achieves fewer parameters (1.67 M) and FLOPs (6.97 G) across all ANN and SNN methods, which demonstrates superior performance in computation efficiency and process speed, achieving a promising 114.29 FPS. 

\begin{figure*}[!t]
    \centering 
\includegraphics[width=0.85\linewidth]{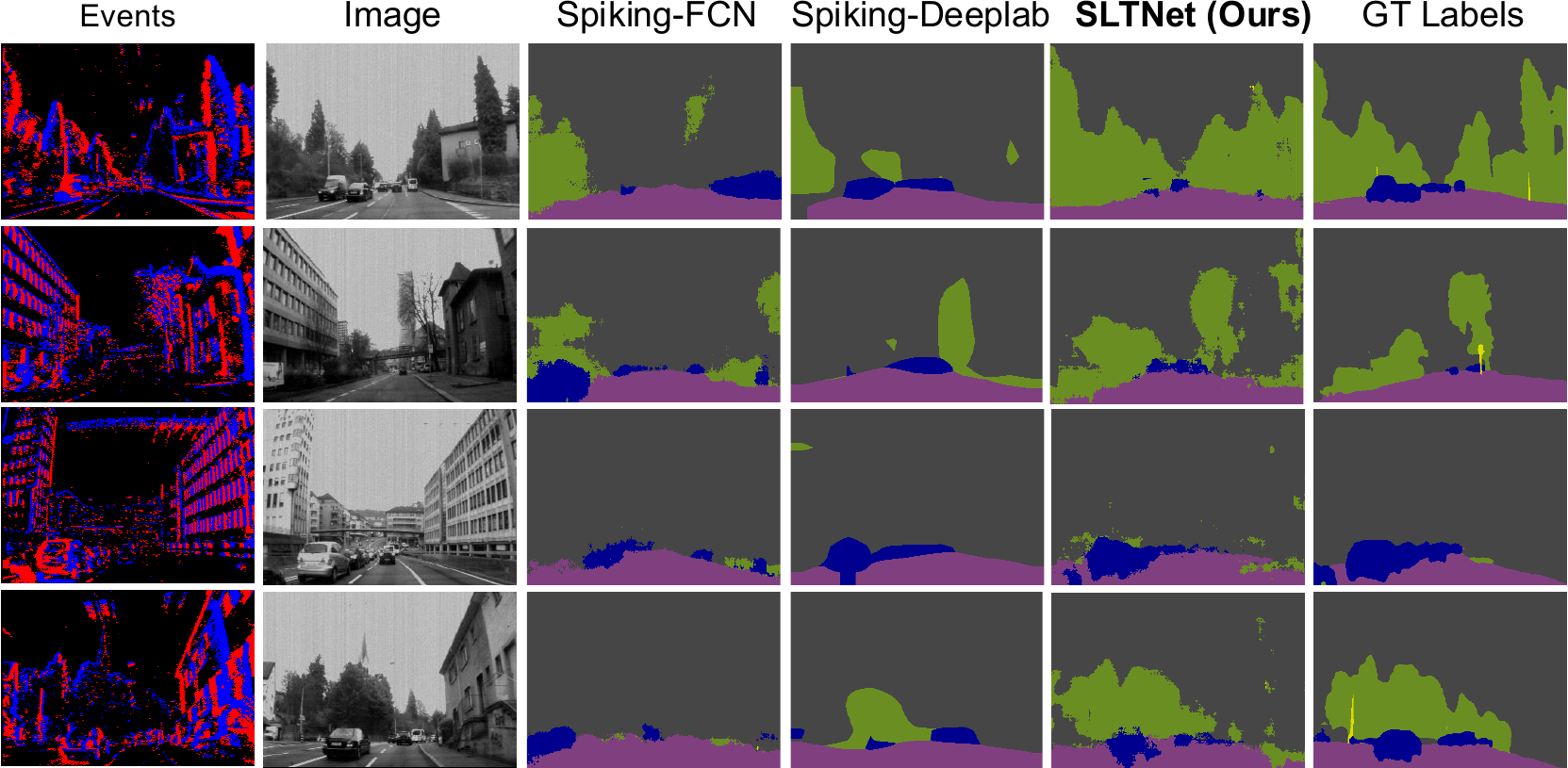}	
    \caption{Qualitative segmentation results on the DDD17 dataset. The left is event frame and its corresponding image frame. The rightmost is
ground truth.} 
    \label{Fig:ddd17_visual}
\end{figure*}

\begin{figure*}[!t]
    \centering 
\includegraphics[width=0.85\linewidth]{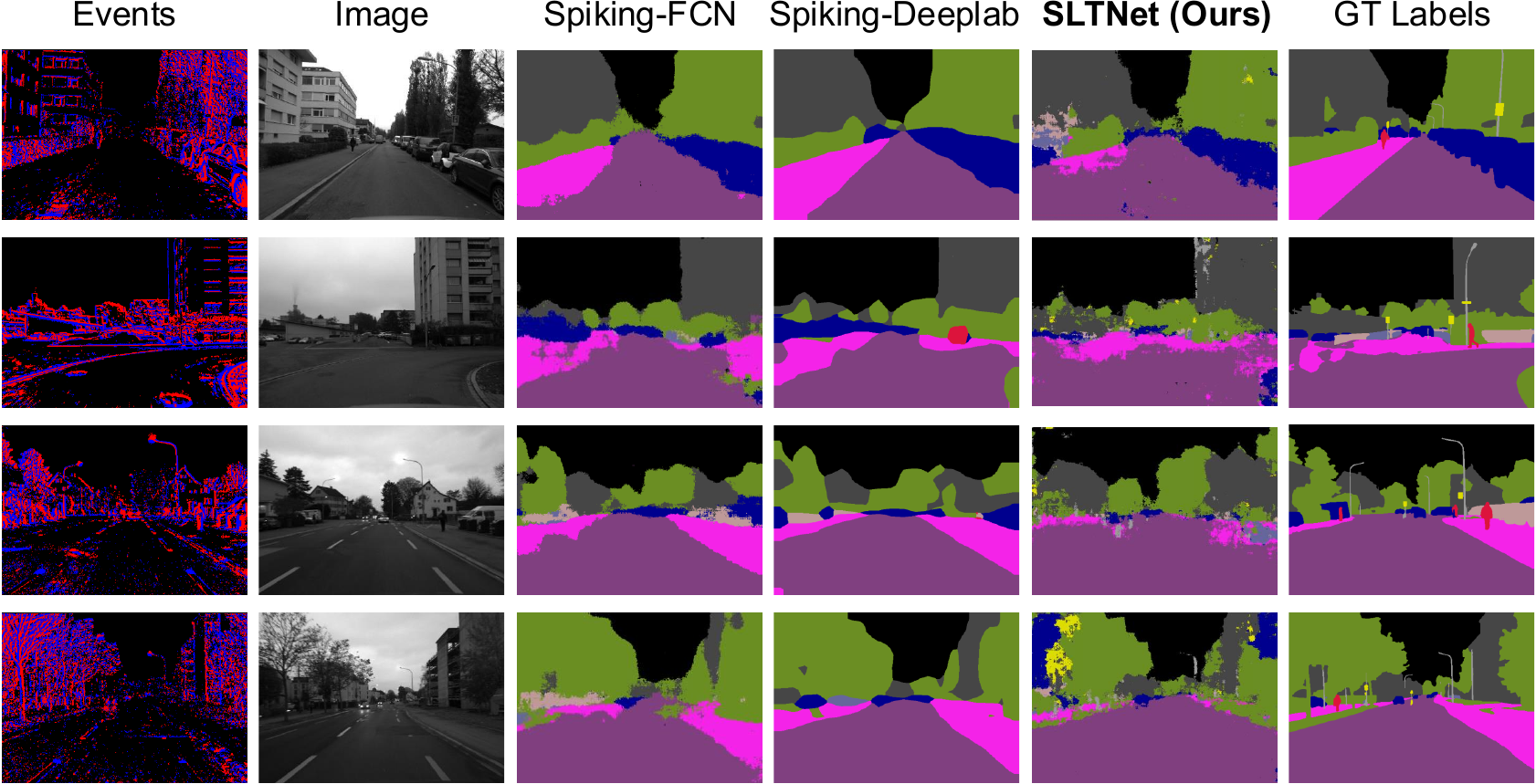}	
    \caption{Segmentation results on the DSEC-Semantic dataset. The left are event frames and corresponding image frames. The rightmost are ground truths.} 
    \label{Fig:dsec_visual}
\end{figure*}

We also present qualitative results in Fig.~\ref{Fig:ddd17_visual} 
and Fig.~\ref{Fig:dsec_visual} 
to provide a more intuitive comparison of different models. It clearly shows that our proposed SLTNet achieves fine-grained segmentation compared to competition models, demonstrating the robustness and effectiveness of our model across different datasets.  


\subsection{Power and Computation Cost Analysis}

Floating-point operations (FLOPs) are widely recognized for evaluating the computational load of neural networks. In ANNs, FLOPs almost come from floating-point matrix multiplication and accumulation (MAC). Since the event-driven characteristics of SNNs, the binary spikes transmitted within them result in almost all FLOPs being accumulated (ACC). Combining the timestep $T=1$ and spiking firing rate $R=0.5$, we can estimate the energy consumption of SNNs:
\begin{equation}
    E = E_{MAC}\times FL_{1} + E_{ACC}\times FL_{2}\times T\times R,
\end{equation}
where $FL$ can be FLOPs of the convolution or linear layer.

Tab.~\ref{tab:energy_cost} compares energy consumption with other models. To maintain consistency with previous work \cite{kim2022beyond}, we set $E_{MAC}=4.6pJ$ and $E_{ACC}=0.9pJ$ during calculation. 
Compared to ANN-based models, our proposed SLTNet excels in energy efficiency. Specifically, Ev-SegNet~\cite{evsegent}, Evdistill~\cite{evdistill}, and ESS~\cite{ess} consume 30.20$\times$, 96.31$\times$, and 37.90$\times$ more energy than our model, respectively. {Although Spiking-DeepLab, Spiking-FCN, and SCGNet also utilize spiking architectures, their straightforward components and design result in 5.48$\times$, 12.00$\times$, and 4.58$\times$ higher power consumption, respectively.}

\subsection{Ablation Studies}

\begin{table}[!t]
    \centering
    \caption{Energy consumption of different ANN/SNN models on DDD17 dataset.}
    \resizebox{0.48\textwidth}{!}
    {
    \begin{tabular}{lcccc}
        \toprule 
        Model & Type & MAC & ACC & Energy ($mJ$) $\downarrow$  \\
        \midrule
        Ev-SegNet & ANN & 9322 M & 0 M & {42.88  (30.20$\times$)}  \\
        Evdistill & ANN & 29730 M & 0 M &  { 136.76  (96.31$\times$)}  \\
        ESS & ANN & 11700 M & 0 M &    {53.82  (37.90$\times$)}      \\
        \midrule
        Spiking-DeepLab & SNN & 1435 M & 2617 M & 7.78 (5.48$\times$)  \\
        Spiking-FCN & SNN & 264 M & 35176 M & 17.04 (12.00$\times$)  \\
        SCGNet & SNN & 140 M & 6500 M & 6.50${\textcolor{blue}{^*}}$ ($4.58\times$)\\
        \textbf{SLTNet (Ours)} & \textbf{SNN} & \textbf{131 M} & \textbf{1830 M} & { \textbf{1.42 (1.00 $\times$) }}\\
        \bottomrule     
    \end{tabular}
    }
    
    \textbf{*}: The energy consumption reported in SCGNet's~\cite{SCGNet} original paper is 6.50 $mJ$. For consistency, we present this value without modification.\\

    \label{tab:energy_cost}
\end{table}

\subsubsection{Spike-Driven Multi-Head Self-Attention}

\begin{table}[!t]
    \centering
    \caption{Ablation studies of Spike-driven Multi-head Self-attention Method on DDD17.}
    \resizebox{0.35\textwidth}{!}
    {
    \begin{tabular}{cc}
        \toprule 
        Self-attention & mIoU (\%)\\
        \midrule
        SDSA2\cite{spikedrivenv2} & 50.77 (-1.16)\\
        SDSA3\cite{spikedrivenv2} & 51.00 (-0.93)\\
        Spike-driven Transformer\cite{spikedriven}& 49.70 (-2.23)\\ 
        Spikformer\cite{spikformer} & 48.74 (-3.19)\\
        \textbf{SDMSA (Ours)} & \textbf{51.93}\\ 
        \bottomrule     
    \end{tabular}
    }
    \label{tab:transformer}
\end{table}

In SNNs, the query, key, and value in self-attention are represented as binary spikes rather than continuous floating-point values. Therefore, traditional self-attention modules designed for ANNs are not directly applicable to SNNs. Developing an efficient self-attention mechanism specifically optimized for SNNs remains a challenge.
In Tab.~\ref{tab:transformer}, we present a performance comparison between our SDMSA attention module and other spike-driven attention/transformer modules, described in Eq. \ref{eq:SDMSA}. The results show that our proposed SDMSA outperforms other methods by at least 1.16\% mIoU when evaluated by the same model. Besides, the proposed SDMSA module is 2.23\% and 3.19\%  higher than the recently proposed spike-driven transformer~\cite{spikedriven} and Spikformer~\cite{spikformer}.

\subsubsection{Shortcut and Architecture}

\begin{table}[!t]
    \centering
    \caption{Ablation studies of SLTNet on DDD17.}
    \resizebox{0.45\textwidth}{!}
    {
    \begin{tabular}{cccccc}
        \toprule 
        Ablation & Module & Params. (M) & mIoU (\%)\\
        \midrule
        \multirow{2}*{Shortcut} & MS $\rightarrow$ VS & 0.41 & 49.03 (-2.90) \\
        ~ & MS $\rightarrow$ SEW & 0.41 & 32.55 (-19.38)\\
        \midrule
        \multirow{3}*{Architrcture} & SLTNet w/o FE  & 0.41 & 49.97 (-1.96)\\
        ~ & SLTNet w/o STB  & 0.41 & 48.47 (-3.46)\\
        ~ & SLTNet w/o Fusion & 0.41 & 49.66 (-2.27)  \\
        \midrule
        \textbf{Ours} & \textbf{SLTNet} & 0.41 & \textbf{51.93} \\
        \bottomrule     
    \end{tabular}
    }
    \label{tab:Spike-LEVSegNet}
\end{table}

The ablation parts on the network architecture have almost no impact on the number of parameters, effectively eliminating the effect of parameter space on performance.

\textbf{Shortcut.} There are three shortcuts in SNNs, where we use the membrane shortcut (MS) that connects the float membrane potential in different layers. The other two are the Spike-Element-Wise (SEW)~\cite{sew_shortcut} Shortcut, which fuses spikes across layers, and the Vanilla Shortcut (VS), which directly connects spikes to membrane potential. As shown in Tab.~\ref{tab:Spike-LEVSegNet}, in our architecture, our MS performs the best, with VS reducing the segmentation accuracy by 2.90\% and SEW shortcut by 19.38\%. The poor performance of the SEW shortcut indicates that using spikes alone without retaining information about the membrane potential can significantly impair performance on tasks requiring detailed information. 

\textbf{Architecture.} When removing the FE module in the skip connection, the mIoU is reduced by 1.96\%, demonstrating its effectiveness in enhancing feature expression. Furthermore, we change the network to fully SCBs without STBs. Performance dropped by 3.46\% significantly, which proves that transformer and convolution blocks can achieve complementarity. Then, we remove the skip connection from stages 1-3 as described in Sec.~\ref{encoder-decoder}. It leads to a performance loss of 2.27\%, showing that the feature map from the early stage can effectively complement segmentation information.

\begin{table}[!t]
    \centering
    \caption{Ablation studies of Loss Design.}
    \resizebox{0.4\textwidth}{!}
    {
    \begin{tabular}{ccc}
        \toprule 
        OHEM Loss ($l_1$) & Early Loss ($l_2$) & mIoU (\%)   \\
        \midrule
        $\times$ & $\times$ & 47.43 (-4.50) \\
        $\times$ & $\checkmark$ & 48.46 (-3.47)  \\
        $\checkmark$ & $\times$ & 49.63 (-2.30)  \\
        $\checkmark$ & $\checkmark$ & \textbf{51.93 (Ours)} \\
        \bottomrule     
    \end{tabular}
    }
    \label{tab:loss}
\end{table}

\subsubsection{Loss Design}
To verify the effectiveness of our loss design, we conduct four groups of ablation experiments on different combinations of two losses in Tab.~\ref{tab:loss}. The results show that using conventional cross-entropy loss (no OHEM and Early loss) leads to a huge performance drop, especially a 4.50\% decrease in mIoU. Notably, the optimization effect of OHEM loss becomes apparent with the help of Early loss, which is 3.47\% and 2.30\% higher than employing OHEM loss alone and early loss alone, respectively.


\section{Conclusion}
In this work, we tackle the real-world challenges of traditional semantic segmentation by designing a lightweight single-branch architecture and integrating it into the spike-based event segmentation paradigm. Specifically, we design an SLTNet that is built on novel SCBs and STBs basic blocks, where SCBs are designed for efficient feature extraction, while STBs are proposed to achieve long-range contextual feature interaction. Extensive experimental results show that our work outperforms SOTA SNN-based methods on both DDD17 and DSEC-Semantic datasets. SLTNet also achieves 4.58$\times$ more energy-saving, and 114 FPS faster inference speed. 
Given the accuracy gap between ANN-based methods and SLTNet, future work will focus on developing a more advanced SNN architecture for event processing and extending our approach to other domains, such as autonomous driving and robotic localization.







\bibliographystyle{IEEEtran}
\bibliography{ref}

\begin{thebibliography}{10}
\providecommand{\url}[1]{#1}
\csname url@samestyle\endcsname
\providecommand{\newblock}{\relax}
\providecommand{\bibinfo}[2]{#2}
\providecommand{\BIBentrySTDinterwordspacing}{\spaceskip=0pt\relax}
\providecommand{\BIBentryALTinterwordstretchfactor}{4}
\providecommand{\BIBentryALTinterwordspacing}{\spaceskip=\fontdimen2\font plus
\BIBentryALTinterwordstretchfactor\fontdimen3\font minus \fontdimen4\font\relax}
\providecommand{\BIBforeignlanguage}[2]{{%
\expandafter\ifx\csname l@#1\endcsname\relax
\typeout{** WARNING: IEEEtran.bst: No hyphenation pattern has been}%
\typeout{** loaded for the language `#1'. Using the pattern for}%
\typeout{** the default language instead.}%
\else
\language=\csname l@#1\endcsname
\fi
#2}}
\providecommand{\BIBdecl}{\relax}
\BIBdecl

\bibitem{ss}
A.~Garcia-Garcia, S.~Orts-Escolano, S.~Oprea, V.~Villena-Martinez, and J.~Garcia-Rodriguez, ``A review on deep learning techniques applied to semantic segmentation,'' \emph{arXiv preprint arXiv:1704.06857}, 2017.

\bibitem{driving}
N.~Theisen, R.~Bartsch, D.~Paulus, and P.~Neubert, ``Hs3-bench: A benchmark and strong baseline for hyperspectral semantic segmentation in driving scenarios,'' in \emph{2024 IEEE/RSJ International Conference on Intelligent Robots and Systems (IROS)}, 2024, pp. 5895--5901.

\bibitem{long_scene_segmentation}
X.~Long, H.~Zhao, C.~Chen, F.~Gu, and Q.~Gu, ``A novel wide-area multiobject detection system with high-probability region searching,'' in \emph{2024 IEEE International Conference on Robotics and Automation (ICRA)}, 2024, pp. 18\,316--18\,322.

\bibitem{Ni_Conext_seg}
Z.~Ni, X.~Chen, Y.~Zhai, Y.~Tang, and Y.~Wang, ``Context-guided spatial feature reconstruction for efficient semantic segmentation,'' in \emph{European Conference on Computer Vision}.\hskip 1em plus 0.5em minus 0.4em\relax Springer, 2024, pp. 239--255.

\bibitem{Ni_ssaseg}
X.~Ma, Z.-L. Ni, and X.~Chen, ``{SSA}-seg: Semantic and spatial adaptive pixel-level classifier for semantic segmentation,'' in \emph{The Thirty-eighth Annual Conference on Neural Information Processing Systems}, 2024.

\bibitem{event_research}
M.~Ikura, C.~L. Gentil, M.~G. Müller, F.~Schuler, A.~Yamashita, and W.~Stürzl, ``Rate: Real-time asynchronous feature tracking with event cameras,'' in \emph{2024 IEEE/RSJ International Conference on Intelligent Robots and Systems (IROS)}, 2024, pp. 11\,662--11\,669.

\bibitem{Gu_eventdrop}
F.~Gu, W.~Sng, X.~Hu, and F.~Yu, ``Eventdrop: Data augmentation for event-based learning,'' in \emph{Proceedings of the Thirtieth International Joint Conference on Artificial Intelligence, {IJCAI-21}}, 2021, pp. 700--707.

\bibitem{spike-brgnet}
X.~Long, X.~Zhu, F.~Guo, C.~Chen, X.~Zhu, F.~Gu, S.~Yuan, and C.~Zhang, ``Spike-brgnet: Efficient and accurate event-based semantic segmentation with boundary region-guided spiking neural networks,'' \emph{IEEE Transactions on Circuits and Systems for Video Technology}, pp. 1--1, 2024.

\bibitem{evsegent}
I.~Alonso and A.~C. Murillo, ``Ev-segnet: Semantic segmentation for event-based cameras,'' in \emph{Proceedings of the IEEE/CVF Conference on Computer Vision and Pattern Recognition Workshops}, 2019, pp. 0--0.

\bibitem{dual}
L.~Wang, Y.~Chae, and K.-J. Yoon, ``Dual transfer learning for event-based end-task prediction via pluggable event to image translation,'' in \emph{Proceedings of the IEEE/CVF International Conference on Computer Vision}, 2021, pp. 2135--2145.

\bibitem{evdistill}
L.~Wang, Y.~Chae, S.-H. Yoon, T.-K. Kim, and K.-J. Yoon, ``Evdistill: Asynchronous events to end-task learning via bidirectional reconstruction-guided cross-modal knowledge distillation,'' in \emph{Proceedings of the IEEE/CVF Conference on Computer Vision and Pattern Recognition}, 2021, pp. 608--619.

\bibitem{ess}
Z.~Sun, N.~Messikommer, D.~Gehrig, and D.~Scaramuzza, ``Ess: Learning event-based semantic segmentation from still images,'' in \emph{European Conference on Computer Vision}.\hskip 1em plus 0.5em minus 0.4em\relax Springer, 2022, pp. 341--357.

\bibitem{yao_sam_segmentation}
B.~Yao, Y.~Deng, Y.~Liu, H.~Chen, Y.~Li, and Z.~Yang, ``Sam-event-adapter: Adapting segment anything model for event-rgb semantic segmentation,'' in \emph{2024 IEEE International Conference on Robotics and Automation (ICRA)}, 2024, pp. 9093--9100.

\bibitem{snn}
S.~Ghosh-Dastidar and H.~Adeli, ``Spiking neural networks,'' \emph{International journal of neural systems}, vol.~19, no.~04, pp. 295--308, 2009.

\bibitem{kim2022beyond}
Y.~Kim, J.~Chough, and P.~Panda, ``Beyond classification: Directly training spiking neural networks for semantic segmentation,'' \emph{Neuromorphic Computing and Engineering}, vol.~2, no.~4, p. 044015, 2022.

\bibitem{evsegsnn}
D.~Hareb and J.~Martinet, ``Evsegsnn: Neuromorphic semantic segmentation for event data,'' in \emph{2024 International Joint Conference on Neural Networks (IJCNN)}, 2024, pp. 1--8.

\bibitem{EventAug_tcds}
F.~Gu, J.~Dou, M.~Li, X.~Long, S.~Guo, C.~Chen, K.~Liu, X.~Jiao, and R.~Li, ``Eventaugment: Learning augmentation policies from asynchronous event-based data,'' \emph{IEEE Transactions on Cognitive and Developmental Systems}, vol.~16, no.~4, pp. 1521--1532, 2024.

\bibitem{mfs-efs}
Q.~Su, W.~He, X.~Wei, B.~Xu, and G.~Li, ``Multi-scale full spike pattern for semantic segmentation,'' \emph{Neural Networks}, vol. 176, p. 106330, 2024.

\bibitem{SCGNet}
H.~Zhang, X.~Fan, and Y.~Zhang, ``Energy-efficient spiking segmenter for frame and event-based images,'' \emph{Biomimetics}, vol.~8, no.~4, 2023.

\bibitem{evaf}
Y.~Guo, Y.~Chen, L.~Zhang, X.~Liu, Y.~Wang, X.~Huang, and Z.~Ma, ``Im-loss: information maximization loss for spiking neural networks,'' \emph{Advances in Neural Information Processing Systems}, vol.~35, pp. 156--166, 2022.

\bibitem{resnet}
K.~He, X.~Zhang, S.~Ren, and J.~Sun, ``Deep residual learning for image recognition,'' \emph{2016 IEEE Conference on Computer Vision and Pattern Recognition (CVPR)}, pp. 770--778, 2015.

\bibitem{CA}
J.~Hu, L.~Shen, and G.~Sun, ``Squeeze-and-excitation networks,'' in \emph{2018 IEEE/CVF Conference on Computer Vision and Pattern Recognition}, 2018, pp. 7132--7141.

\bibitem{yao2024spikedrivenv2}
\BIBentryALTinterwordspacing
M.~Yao, J.~Hu, T.~Hu, Y.~Xu, Z.~Zhou, Y.~Tian, B.~XU, and G.~Li, ``Spike-driven transformer v2: Meta spiking neural network architecture inspiring the design of next-generation neuromorphic chips,'' in \emph{The Twelfth International Conference on Learning Representations}, 2024. [Online]. Available: \url{https://openreview.net/forum?id=1SIBN5Xyw7}
\BIBentrySTDinterwordspacing

\bibitem{repconv}
\BIBentryALTinterwordspacing
S.~Deng and S.~Gu, ``Optimal conversion of conventional artificial neural networks to spiking neural networks,'' in \emph{International Conference on Learning Representations}, 2021. [Online]. Available: \url{https://openreview.net/forum?id=FZ1oTwcXchK}
\BIBentrySTDinterwordspacing

\bibitem{sew_shortcut}
W.~Fang, Z.~Yu, Y.~Chen, T.~Huang, T.~Masquelier, and Y.~Tian, ``Deep residual learning in spiking neural networks,'' \emph{Advances in Neural Information Processing Systems}, vol.~34, pp. 21\,056--21\,069, 2021.

\bibitem{OHEM}
A.~Shrivastava, A.~Gupta, and R.~Girshick, ``Training region-based object detectors with online hard example mining,'' in \emph{2016 IEEE Conference on Computer Vision and Pattern Recognition (CVPR)}, 2016, pp. 761--769.

\bibitem{ddd17}
J.~Binas, D.~Neil, S.-C. Liu, and T.~Delbruck, ``Ddd17: End-to-end davis driving dataset,'' \emph{arXiv preprint arXiv:1711.01458}, 2017.

\bibitem{dsec}
Z.~Sun, N.~Messikommer, D.~Gehrig, and D.~Scaramuzza, ``Ess: Learning event-based semantic segmentation from still images,'' in \emph{European Conference on Computer Vision}.\hskip 1em plus 0.5em minus 0.4em\relax Springer, 2022, pp. 341--357.

\bibitem{spikedrivenv2}
M.~Yao, J.~Hu, T.~Hu, Y.~Xu, Z.~Zhou, Y.~Tian, B.~XU, and G.~Li, ``Spike-driven transformer v2: Meta spiking neural network architecture inspiring the design of next-generation neuromorphic chips,'' in \emph{The Twelfth International Conference on Learning Representations}, 2024.

\bibitem{spikedriven}
M.~Yao, J.~Hu, Z.~Zhou, L.~Yuan, Y.~Tian, B.~XU, and G.~Li, ``Spike-driven transformer,'' in \emph{Thirty-seventh Conference on Neural Information Processing Systems}, 2023.

\bibitem{spikformer}
Z.~Zhou, Y.~Zhu, C.~He, Y.~Wang, S.~YAN, Y.~Tian, and L.~Yuan, ``Spikformer: When spiking neural network meets transformer,'' in \emph{The Eleventh International Conference on Learning Representations}, 2023.

\end{thebibliography}

\end{document}